\documentclass[sigconf]{acmart}
\AtBeginDocument{%
  }

\usepackage{booktabs} 
\usepackage{multirow} 
\usepackage{colortbl} 
\usepackage{rotating} 
\usepackage{url}

\usepackage{caption}
\usepackage{graphicx}
\usepackage{float} 
\usepackage{subcaption}
\usepackage{CJKutf8}

\copyrightyear{2025}
\acmYear{2025}
\setcopyright{cc}
\setcctype{by}
\acmConference[CIKM '25] {Proceedings of the 34th ACM International Conference on Information and Knowledge Management}{November 10--14, 2025}{Seoul, Republic of Korea.}
\acmBooktitle{Proceedings of the 34th ACM International Conference on Information and Knowledge Management (CIKM '25), November 10--14, 2025, Seoul, Republic of Korea}
\acmISBN{979-8-4007-2040-6/2025/11}
\acmDOI{10.1145/3746252.3761428}

\begin{document}

\title[TCM-HEDPR: A Hierarchical Structure-Enhanced Personalized HPR Model for TCM Formulas]{A Hierarchical Structure-Enhanced Personalized Recommendation Model for Traditional Chinese Medicine Formulas Based on KG Diffusion Guidance}


\author{Chaobo Zhang}
\orcid{0009-0005-2717-9172}
\affiliation{%
  \institution{Heilongjiang University}
  \city{Harbin}
  \country{China}
}
\email{2231974@s.hlju.edu.cn}

\author{Long Tan}
\authornote{Corresponding author.}
\orcid{0000-0001-8444-5362}
\affiliation{%
  \institution{Heilongjiang University}
  \city{Harbin}
  \country{China}
}
\email{tanlong@hlju.edu.cn}








\renewcommand{\shortauthors}{Chaobo Zhang and Long Tan}

\begin{abstract}
Artificial intelligence (AI) technology plays a crucial role in recommending prescriptions for traditional Chinese medicine (TCM). 
Previous studies have made significant progress by focusing on the symptom-herb relationship in prescriptions. However, several limitations hinder model performance: (i) Insufficient attention to patient-personalized information such as age, BMI, and medical history, which hampers accurate identification of syndrome and reduces efficacy. (ii) The typical long-tailed distribution of herb data introduces training biases and affects generalization ability. (iii) The oversight of the ‘monarch, minister, assistant and envoy’ compatibility among herbs increases the risk of toxicity or side effects, opposing the ‘treatment based on syndrome
differentiation’ principle in clinical TCM. Therefore, we propose a novel hierarchical structure-enhanced personalized recommendation model for TCM formulas based on knowledge graph (KG) diffusion guidance, namely TCM-HEDPR. Specifically, we pre-train symptom representations using patient-personalized prompt sequences and apply prompt-oriented contrastive learning (CL) for data augmentation. Furthermore, we employ a KG-guided homogeneous graph diffusion method integrated with a self-attention mechanism to globally capture the non-linear symptom-herb relationship. Lastly, we design a heterogeneous graph hierarchical network to integrate herbal dispensing relationships with implicit syndromes, guiding the prescription generation process at a fine-grained level and mitigating the long-tailed herb data distribution problem. Extensive experiments on two public datasets and one clinical dataset demonstrate the effectiveness of TCM-HEDPR. In addition, we incorporate insights from modern medicine and network pharmacology to evaluate the recommended prescriptions comprehensively. It can provide a new paradigm for the recommendation of modern TCM.
\end{abstract}




\maketitle

\section{Introduction}

Traditional Chinese Medicine (TCM), as a treasure of Chinese culture, embodies the profound legacy and unique insights of millennia-old practices, especially plays a vital role in treating COVID-19 and other severe illnesses \cite{b1,b2}. TCM practitioners diagnose and treat patients primarily through four methods: observation (wàng), smelling (wén), questioning (wèn), and pulse-taking (qiè) to identify symptom characteristics. These are then combined with the patient's personal attributes and medical history to give a suitable combination of herbs as a prescription \cite{b3}.

Studies have indicated that personalized patient information, such as age, gender, BMI, and medical history, is integral to herb prescribing and an essential factor in clinical herbalist practice \cite{b4,b5,b6}. In the diagnostic case illustrated in Fig. \ref{fig_1}, the patient presented with symptoms like chest tightness and palpitations. The TCM doctor considered the patient's history of diabetes and their BMI ($weight/height^2 = 31.6$, indicating obesity), concluding that the primary syndrome was "phlegm-dampness-Qi stagnation syndrome." This diagnosis was preferred over other herbal prescription recommendation (HPR) methods that incorrectly identified the syndrome as "anemia of Qi and blood" based solely on the symptoms. Consequently, Semen Trichosanthis was chosen as the primary 'monarch' herb for addressing chest paralysis and phlegm, while Allium Macrostemon and Pinellia Ternata served as 'minister' herbs to enhance the effects of the 'monarch' herb by expelling phlegm and dispersing lumps (\begin{CJK*}{UTF8}{gbsn}祛痰散结\end{CJK*}). Additionally, Hawthorn Meat and Hedge Thorn were selected as 'assistant \& envoy' herbs to invigorate Qi and alleviate Qi stagnation (\begin{CJK*}{UTF8}{gbsn}行气除满\end{CJK*}), as well as to lower blood sugar levels.

Currently, techniques such as AI and machine learning have been devoted to simulating the (HPR) process. For example, SMGCN \cite{b7} utilizes a bipartite graph convolution approach to learn symptom-herb relationships. KG-ASMGCN \cite{b8} improves an attention mechanism to SMGCN and extracts fixed-dimensional embeddings in KG as a supplement to additional TCM knowledge. KDHR \cite{b9} applies a multi-graph convolution approach to learn higher-order representations such as symptom-herb and herb-herb. SMRGAT \cite{b10} incorporates the multi-graph residual attention networks for the fusion of semantic knowledge of herbs into HPR. PresRecST \cite{b11} draws upon the attention weights and KG in order to simulate the progressive diagnosis and treatment process of TCM. LAMGCN \cite{b12} performs the LSTM technique to learn a priori about herbs by disambiguating canonical texts from TCM. TCMRGCL \cite{b13} carries out the contrastive pre-training and hierarchical structure to capture the symptom-herb relationship. SDPR \cite{b14} proposes a four-part graph strategy to deal with the relationship between symptoms, herbs, etc. However, the above models do not really achieve individualized as well as progressive simulation of the clinical TCM diagnosis and treatment process for patients and are less convincing and interpretable.

At this stage, HPR faces the following four main challenges: (1) Scarcity of patient personalized data. (2) The frequency of herbs data shows a long-tailed distribution[imbalanced labels (herbs)]. (3) Incorporating the implicit syndrome induction process. (4) Capturing complex herb pairing relationships.

To address the above problems, this paper proposes a novel HPR model, namely TCM-HEDPR, which includes five main modules: (1) Patient individualized feature pre-embedding module (PEPP). (2) Diffusion-guided symptom-herb representation learning module (DMSH). (3) Syndrome-aware prediction module (SYN). (4) Heterogeneous graph-enhanced hierarchical structured network of herbs (HGSN). (5) HPR module (PR). To summarize, TCM-HEDPR's contribution can be concluded as follows:

\begin{figure}[!t]
\centering
\includegraphics[width=0.45\textwidth, angle=0,scale=1.0]{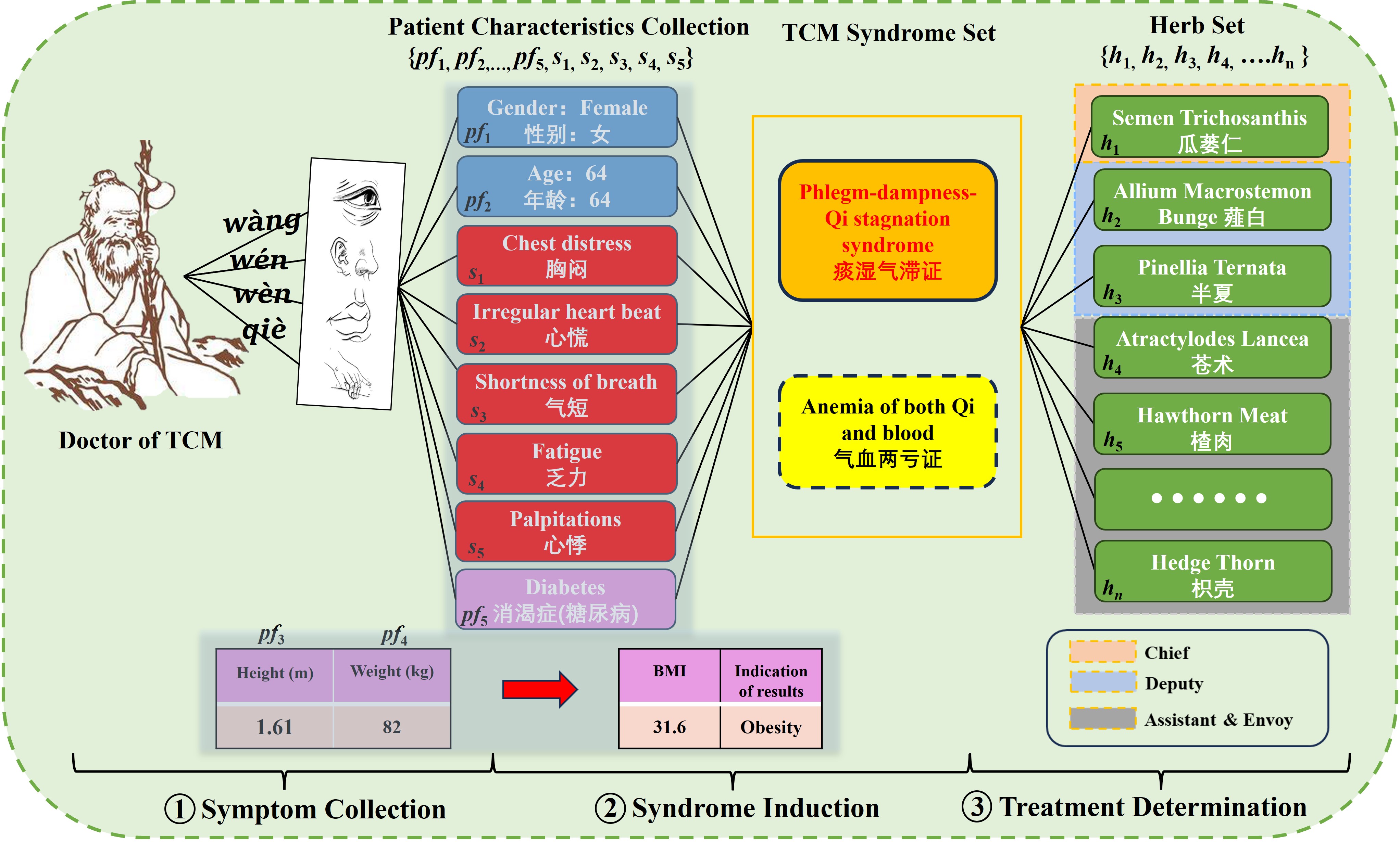}
\caption{Diagram of the diagnostic and therapeutic process in traditional Chinese medicine (TCM).}
\label{fig_1}
\end{figure}

\begin{itemize}
\item We propose an innovative hierarchical heterogeneous graph network to capture the 'monarch, minister, assistant and envoy' relationship of herbs and incorporate it into the implicit syndrome summarization process. To the best of our knowledge, we are pioneers in integrating it into the HPR framework.
\item  We pre-embed patient-personalized information in the form of prompt sequences and augment it base on CL while capturing the multivariate symptom-herb relationship using a diffusion probability model and interpreting the results of the herbs' predictions by means of a network pharmacology approach.
\item  We construct a large-scale TCM knowledge graph and incorporate it into a diffusion probabilistic learning process, which enriches the relationships between entities like symptoms and herbs while effectively modeling the semantic and synergistic interactions among knowledge-aware herbs through the KG and diffusion model integration.
\item  Experiments were conducted on two public datasets and one clinical dataset, demonstrate that TCM-HEDPR outperforms current HPR methods, offering valuable support for clinical diagnosis and decision-making in TCM.
\end{itemize}

\section{Math and methods}

\subsection{Problem definition}

HPR aims to summarize the syndrome based on the patient's personal information (e.g. gender, age, medical history, etc.) in conjunction with the symptom characteristics and to give an appropriate combination of herbs as a prescription, and we formulated the question as follows.

Give a patient set \textit{U}=$\{u_1,u_2,\ldots,u_z\}$,  where $u_i$ denotes patient \textit{i}. Each patient has a personal feature set \textit{PF}=$\{{{pf}_1,{pf}_2,\ldots,{pf}_t}\}$, with ${pf}_i$ representing specific characteristics. The \textit{PF} set contains at least 4 types of features (${pf}_1-{pf}_4$) : gender, age, height, and weight. Additional features ${pf}_5$ through ${pf}_t$ may cover the patient's medical history (optional). Meanwhile, the set of herbs \textit{H}=$\{{h_1,h_2,\ldots,h_m}\}$ contains $m$ different types of herbs, and the set of symptoms \textit{S}=$\{{s_1,s_2,\ldots,s_n}\}$ includes $n$ different symptoms. A patient's prescription \textit{P}=$<\left\{{pf}_1,\ldots,{pf}_t\right\},\left\{s_1,\ldots{,s}_n\right\},\left\{h_1,\ldots{,h}_m\right\}\ $ $>=({pf}_{set}$,$h_{set})$, where for each patient $u_i$, there is a corresponding set $h_{set}$ of recommended herbs based on their unique characteristics ${pf}_{set}$ and related symptoms. In addition, we manually constructed the TCM knowledge graph $\mathcal{G}_k=(V,E)$ with the node set $\textit{V}=\{{V_1,V_2,\ldots,V_n}\}$ and the edge set $E\subseteq V\times V$, where $(V_i,V_j)\in E$ denotes the edges from the node $V_i$ to the node $V_j$.

Given a set of patients' symptom information ${pf}_{set}$ and a TCM knowledge graph $\mathcal{G}_k$, the task of this paper is to train a recommendation model $\mathcal{F}({pf}_{set},\mathcal{G}_k,H;\theta)$ to recommend a set of herbs $h_{set}$, where $\theta$ represents the learnable parameters. 

\subsection{TCM knowledge graph}

Currently, since there is no authoritative and publicly certified generic TCM knowledge graph (KG), we use Neo4j \footnote{\url{https://neo4j.com/}} to construct a TCM knowledge graph, namely TCM\_IKG, whose data are mainly derived from the \textit{Encyclopedia of Chinese Medicine}, \textit{Chinese} \textit{Pharmacopoeia (2020 edition)}, the data accumulated by the team, and TCM\_KG \cite{b15}. The fundamental unit of TCM\_IKG is represented by the triplet $<h,r,t>$, where $h$ is the head entity, $t$ is the tail entity, and $r$ is the relationship. We predefined 16 types of entities based on TCM domain knowledge, including herb, prescription, compound, disease, dosage, and efficacy, among others. TCM\_IKG comprises 130,142 entities and 1,145,020 triplets. 

\begin{figure*}[!t]
\centering
\includegraphics[width=0.6\textwidth, angle=0,scale=1.35]{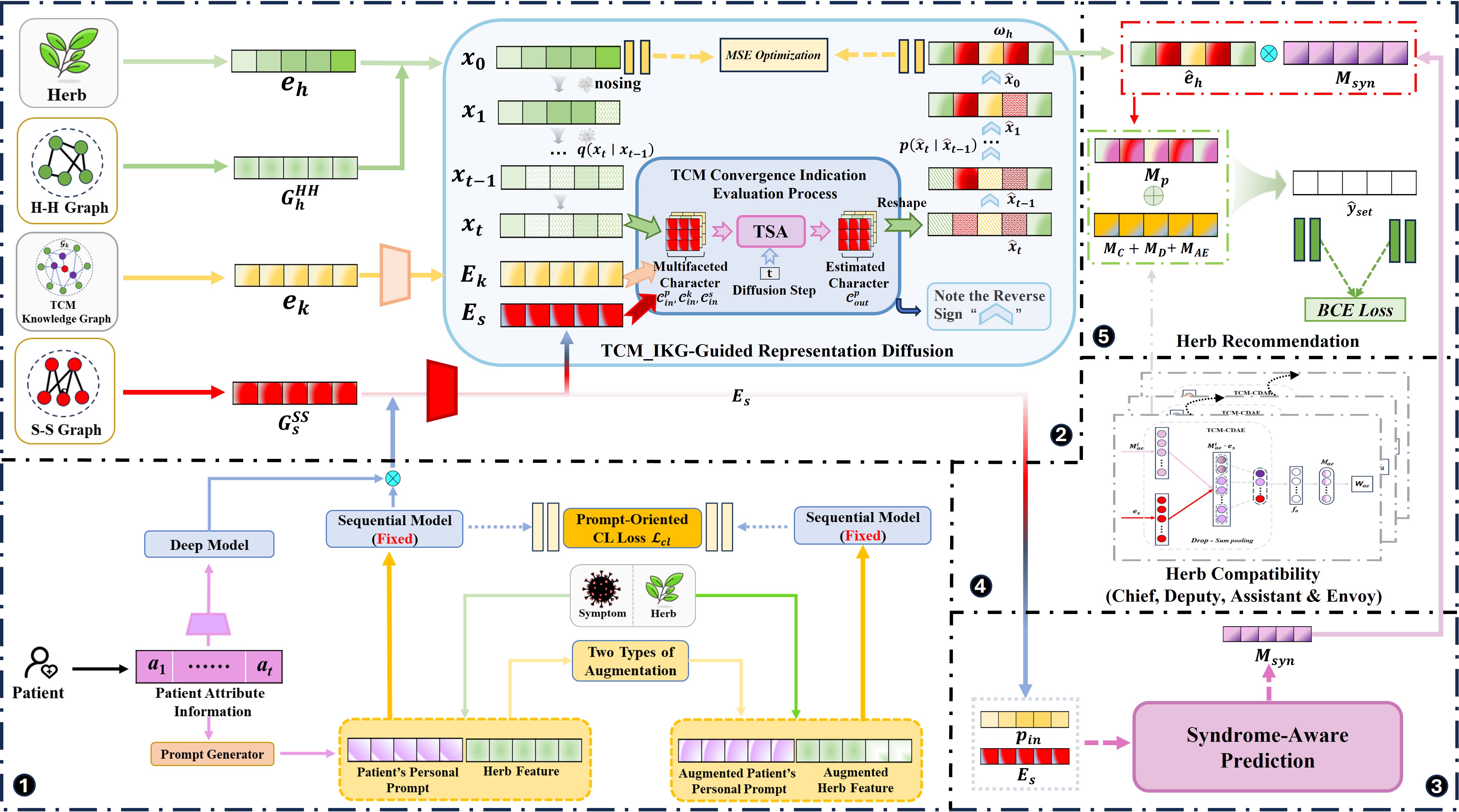}
 \caption{The Overview of TCM-HEDPR. TCM-HEDPR consists of five important modules whose main work: (1) Pre-embedding of patient personalized information; (2) Global modeling of symptom-herb relationships; (3) Incorporation of implicit syndrome induction processes; (4) Fine-grained capturing of symptom-herb and herb-herb relationships combined with the (2) process of mitigating the long-tailed distribution of their data; (5) Personalized HPR process.}
\label{fig_2}
\end{figure*}

\subsection{Overview of the TCM-HEDPR}

The framework of TCM-HEDPR is shown in Fig. \ref{fig_2}. TCM-HEDPR consists of five main modules: (1) The PEPP module, we employ prompt techniques to pre-embed personalized patient attribute sequences and enhance them using CL. (2) The DMSH module, we leverage a diffusion probability model (DM) to learn the relationship between the pre-embedded symptom information and the herb and utilize the TCM\_IKG to supplement the additional TCM knowledge. In addition, we designed a TCM aggregation indicator block (TSA) to enhance the aggregated representation of herb features. (3) The HGSN module, we design a hierarchical structured network to learn the ‘monarch, minister, assistant and envoy’ compatibility relationship between herbs and combine it with the SYN module to predict the symptom preference of herbs, so as to recommend the most appropriate herbs.

\subsection{Pre-training for patient personalization}

Inspired by the prompt fine-tuning technique in the recommendation domain  \cite{b16,b17,b18}, we utilize prompt sequences to pre-train patient-specific symptom and herb relationships.

\subsubsection{TCM symptom prompt definition}

We define the pre-trained model as $f_{seq}(s_1^u, \ldots, s_{|PF|}^u, h_1^u, \ldots, h_{|PF|}^u | \theta)$, and refer to the fine-tuned model as $f_{seq}(s_1^u, \ldots, s_{|PF|}^u, h_1^u, \ldots, h_{|PF|}^u | \hat{\theta})$. The symptom order with patient-specific attribute information is denoted as $f_{seq}(s_1^{pf_1^u, \ldots, pf_T^u}, s_{|PF|}^{pf_1^u, \ldots, pf_T^u}, h_1^u, \ldots, h_{|PF|}^u | \theta, \vartheta)$, where $pf_1^u, \ldots, pf_T^u$ represents patient $u$'s personal attributes, and $\vartheta$ is the parameter for prompt generation.

\subsubsection{TCM symptom prompt pre-training}

To extract the nonlinear complexity of the symptom-herb relationship, we use SASRec \cite{b33} as a pre-training model, which is widely used for prompt training \cite{b17,b34}. For the target herb sequence $H$, we define its $l$-th layer's matrix as $H_u^l$=$\{{h_{h,1}^l,\ldots,h_{h,N}^l}\}$, where $h_{h,i}^l$ denotes the \textit{i}-th behavioral representation of herb $h$ at the $l$-th layer. Then the herb matrix $H_u^{l+1}$ at the $(l+1)$-th layer is represented as:
\begin{equation}
\label{deqn_ex1a_1}
\begin{aligned}
\begin{array}{c}H_{u}^{l+1}=\text { Transformer }^{l}\left(H_{u}^{l}\right)\\\end{array}
\end{aligned}
\end{equation}
\begin{equation}
\begin{aligned}
\label{deqn_ex1a_2}
s_{0}=f_{\textit {seq }}\left(\left\{s_{1}^{u}, \ldots, s_{|P F|}^{u}, h_{1}^{u}, \ldots, h_{|P F|}^{u} \mid \theta\right\}\right)=h_{h, N}^{L}
\end{aligned}
\end{equation}
where $L$ represents the number of Transformer layers and $s_0$ is the learned final symptom, embedded in the subsequent herb during known symptom prediction in pre-training.

\subsubsection{Patient personalization prompt generator}

We define the patient's personalized attributes $x_{pf}=[{pf}_1^s|,\ldots,|{pf}_T^s]$, where ${pf}_i^s$ denotes the $i$ th patient attribute information for symptom $s$. We utilize a simple and efficient MLP as a patient's prompt representation generator \cite{b16,b17}.
\begin{equation}
\label{deqn_ex1a_3}
I^u= \operatorname{MLP}\left(x_{pf}\middle|\vartheta\right)=W_2\sigma\left(W_1x_{pf}+b_1\right)+b_2
\end{equation}
where $W_1$, $W_2$, $b_1$ and $b_2$ are trainable parameters.

\subsubsection{Fine-tuning of symptom prompt}

Based on prefix fine-tuning, we have finalized the formulation of symptom representations by integrating the patient's personalized attribute prompt tokens as prefixes with the corresponding set of herbs. Specifically, we construct the enhanced symptom sequence $\widehat{s}_u=\{{I_1^u,\ldots,I_n^u,h_1^u,\ldots,h_{\left|PF\right|}^u\}}$ using Eq. (\ref{deqn_ex1a_1}) and combine it with the acquired personalized prompt representation $\mathcal{U}_a$ to derive the ultimate symptom representation $\mathcal{U}_s$.
\begin{equation}
\label{deqn_ex1a_4}
\begin{aligned}
\begin{array}{lc}
&\mathcal{U}_s=\mathcal{U}_s^I+\mathcal{U}_a\\
&\mathcal{U}_a={\operatorname{MLP}}_a(x_{pf}|\psi)\\
&\mathcal{U}_s^I=f_{seq}(x_{pf}|\theta,\vartheta)
\end{array}
\end{aligned}
\end{equation}

\subsection{Diffusion-guided symptom-herb representation learning}

\subsubsection{Pre-definition of the TCM\_IKG}

To capture the higher-order semantic relationships between herb and other entities in TCM\_IKG, motivated by KGAT \cite{b19}, we represent ${e}_h^{(k)}$ for the embedding of the herb entity $e_h$ in layer $k$ as follows.
\begin{equation}
\label{deqn_ex1a_5}
\begin{aligned}
e_h^{(k)}={\rm AGGREGATE}^{(k)}(e_h^{(k-1)},\sum_{t\in\mathcal{N}(e_h,r)} w_{(e_h,r,t)}e _t^{(k-1)})
\end{aligned}
\end{equation}
where $w_{\left(e_h,r,t\right)}=softmax\left(\left(W_2e_h+e_r\right)\ast W_2e_t\right)$ denotes the weights. We perform an aggregated representation of the herbal entity feature vector embedding for each layer, i.e., $E^{\left(k\right)}=\left[e_1^{\left(k\right)},\ldots,e_h^{\left(k\right)}\right]$, where $E^{\left(k\right)}=\operatorname{SUM}({NN}_1((E^{\left(k-1\right)}+W_3E^{\left(k-1\right)}a_1^{\left(k\right)};a_3^{\left(k\right)}),{NN}_2((\\E^ {(k-1)}\odot W_3E^{(k-1)}a_2^{(k)});a_4^{(k)}))$, ${NN}_1$ and ${NN}_2$ denote the neural network activation functions. We obtain the final representation denoted as  $E_k={\rm CONCAT}(E^{(0)}, ... ,E^{(k)})$.

\subsubsection{Construction of homogeneous graphs}

To identify the complex interrelationships between symptoms and herbs in prescriptions, we constructed symptom-symptom and herb-herb homogeneity graphs referring to the HPR method \cite{b7,b13}. We denote the symptom and herb representations trained by GAT as $G_s^{SS}$ and $G_h^{HH}$, respectively.

\subsubsection{TCM Knowledge diffusion process}

Inspired by the DM in the recommendation domain \cite{b20,b21,b22}. Our DMSH is similar to other DM methods, which mainly consist of two processes: forward and reverse. Our aim is to model the higher-order knowledge embedded in herb through DM, so as to provide a priori knowledge for processing in subsequent modules.

(1) Forward process: We define $x_0$ as the initialized index embedding $E_h$ of herb $h$. In the forward process, we construct a Markov chain where the Gaussian noise is gradually incorporated into $x_0$ as follows.
\begin{equation}
\label{deqn_ex1a_6}
\begin{aligned}
\begin{array}{c}q\left(x_{1: T} \mid x_{0}\right):=\prod_{t=1}^{T} q\left(x_{t} \mid x_{t-1}\right) \\q\left(x_{t} \mid x_{t-1}\right):=\mathcal{N}\left(x_{t} ; \sqrt{1-\beta_{t}} x_{t-1}, \beta_{t} I\right)\end{array}
\end{aligned}
\end{equation}
where $E_h=\frac{1}{2}(e_h+G_h^{HH})$, \textit{t} denotes the diffusion step and $\beta_t\in\left(0,1\right)$ denotes the scale of added noise. At each step, our forward process is computed as:
\begin{equation}
\label{deqn_ex1a_7}
\begin{aligned}
\begin{array}{l}
q\left(x_{t} \mid x_{0}\right)=\mathcal{N}\left(x_{t} ; \sqrt{\bar{\alpha}_{t}} x_{t},\left(1-\bar{\alpha}_{t}\right) I\right)
\end{array}
\end{aligned}
\end{equation}
where $\epsilon\sim\mathcal{N}(0,I)$, $\alpha_t=1-\beta_t$ and ${\overline{\alpha}}_t=\prod_{t^\prime=1}^{t}\alpha_{t^\prime}$. Then we obtain $x_t=\sqrt{{\overline{\alpha}}_t}x_0+\sqrt{1-{\overline{\alpha}}_t}\epsilon$.

(2) Reverse process: Our reverse process employs a neural network to iteratively recover and learn transitions from $x_t$ to $x_{t-1}$, thereby approximating the initial herb representation $x_0$ by reversing the evolution from $x_t$.
\begin{equation}
\label{deqn_ex1a_8}
\begin{aligned}
\begin{array}{c}
p_{\theta}\left(x_{0: T}\right)=p\left(x_{T}\right) \prod_{t=1}^{T} p_{\theta}\left(x_{t-1} \mid x_{t}\right) \\p_{\theta}\left(x_{t-1} \mid x_{t}\right)=\mathcal{N}\left(x_{t} ; \mu_{\theta}\left(x_{t}, t, E_{h},E_{k},E_{s}\right), \Sigma_{\theta}\left(x_{t}, t\right)\right)
\end{array}
\end{aligned}
\end{equation}
where $\Sigma_\theta(x_t,t)=\sigma_t^2I=\frac{1-{\overline{\alpha}}_{t-1}}{1-{\overline{\alpha}}_t}\beta_tI$ denotes the variance while $E_s=\frac{1}{2}(e_s+G_ s^{SS})$ represents the potential representation of the association with symptom $s$ obtained in the previous step. Meanwhile, this mean $\mu_\theta(x_t,t,E_{h},E_{k},E_{s})$ is calculated as follows.
\begin{equation}
\label{deqn_ex1a_9}
\begin{aligned}
\mu_{\theta}\left(x_{t}, t, E_{h},E_{k},E_{s}\right)=\frac{1}{\sqrt{\alpha_{t}}}\left(x_{t}-\frac{\beta_{t}}{\sqrt{1-\bar{\alpha}_{t}}} f_{\theta}\left(x_{t}, t, E_{h},E_{k},E_{s}\right)\right)
\end{aligned}
\end{equation}
where $f_\theta(\cdot)$ we use a three-layer neural network with feedforward layers.

After $t$ steps, we obtain this predicted value ${\hat{x}}_0$ denoted as $x_p$. In order to preserve more personalized features related to herb while introducing the TCM\_IKG, we use ${\hat{e}}_h = x_0 + \omega_h x_p$ as the final representation for the subsequent HPR task, where $\omega_h$ is the predefined diffusion weight.

\subsubsection{TCM aggregation indication process}

To enable the herb representations to capture global information in multiple entity embedding fusion, we introduce a Transformer \cite{b23} based multiple self-attention mechanism (TSA) after aggregating the features. Specifically, we aggregate the predicted herb representation $x_p$,  $E_k$ and symptom representation $E_s$ as independent channels $\mathcal{C}_{in}^p$,$\mathcal{C}_{in}^k$ and $\mathcal{C}_{in}^s$ of convolutional neural network (CNN) modeling and then input them into the TSA component. At the same time, we add the diffusion step representation vector $t_i$ to each CNN layer along with the above channel matrices. Through the forward propagation of TSA, we obtain the evaluated channel indicators $\mathcal{C}_{out}^p$, $\mathcal{C}_{out}^k$, and $\mathcal{C}_{out}^s$ while recovering $\mathcal{C}_{out}^p$ to the predictive representation ${\hat{x}}_0$. After the above operations, our TCM aggregation indication process can constantly and precisely regulate the estimation of ${\hat{x}}_0$.

\subsubsection{TCM diffusion loss optimization}

Similar to other DM methods, we utilize the mean square error loss (MSE) to represent the optimization process.
\begin{equation}
\label{deqn_ex1a_10}
\begin{aligned}
\mathcal{L}_{T C M \_D M}=E_{x_{0}, x_{t}}\left[\left\|x_{0}-f_{\theta}\left(x_{t}, t, E_{h},E_{k},E_{s}\right)\right\|^{2}\right]
\end{aligned}
\end{equation}

\subsection{Syndrome-aware prediction and herb compatibility}

\subsubsection{Syndrome-aware prediction}

For patients, varying symptoms define the core syndromes, which are closely linked to herbs \cite{b24}. Consequently, we summarize the syndromes using a dot product attention mechanism based on Transformer \cite{b23} with scaling, whose main operation is the introduction of a multi-head self-attention mechanism with $N$ heads. It is represented as follows.
\begin{equation}
\resizebox{1\hsize}{!}{$
\begin{aligned}
\label{deqn_ex1a_11}
M_{syn}={\operatorname{CONCAT}}_{l=0}^N[\operatorname{Attention}(W_Q^{\left(l\right)}\cdot{pf}_{in},W_K^{\left(l\right)}{pf}_{in})W_V^{\left(l\right)}\cdot {pf}_{in}],
\end{aligned}
$}
\end{equation}
\begin{equation}
\begin{aligned}
\label{deqn_ex1a_12}
\begin{array}{l}\operatorname{Attention}\left(W_{Q}^{(l)} \cdot p f_{in }, W_{K}^{(l)} \cdot p f_{in }\right) \\=\operatorname{softmax} \frac{\left(W_{Q}^{(l)} \cdot p f_{i n}\right)^{T}\left(W_{K}^{(l)} \cdot p f_{i n}\right)}{\sqrt{D}}\end{array}
\end{aligned}
\end{equation}
where ${pf}_{in}={pf}_{one-hot}\cdot E_s$,  ${pf}_{one-hot}$ is the input patient personalized symptom vector. $W_Q^{\left(l\right)}$, $W_K^{\left(l\right)}$ and $W_V^{\left(l\right)}$ are the training parameters, while $N$ determines the number of possible syndromes.

\subsubsection{Incorporation of herb compatibility relationship}

\begin{figure}[h]
\centering
\includegraphics[width=0.42\textwidth, angle=0,scale=1.15]{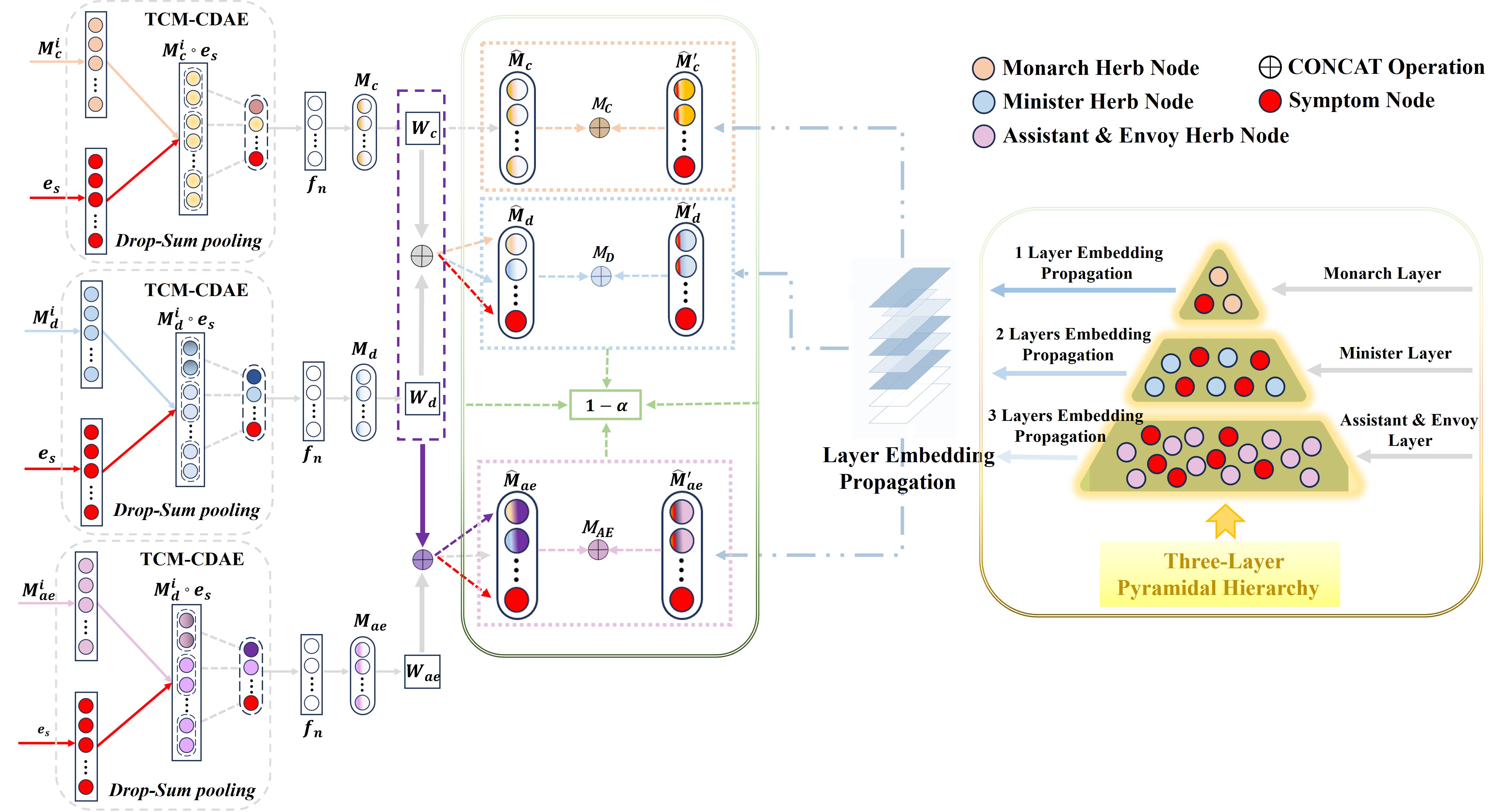}\caption{Hierarchical herb compatibility relationship network.}
\label{fig_3}
\end{figure}

To clearly express and simulate the compatibility relationship between herbs and HPR more realistically, we briefly describe them while designing a hierarchical network to learn these relationships, as illustrated in Fig. \ref{fig_3}.

(1) The 'monarch' herb is indispensable in prescriptions and plays a crucial role in treating the main symptom. (2) The 'minister' herb complements the 'monarch' herb by enhancing the treatment of these primary symptoms. (3) The 'assistant \& envoy' herbs primarily serve as adjuvants; they treat secondary symptoms or mitigate the toxicity of the 'monarch \& minister' herbs \cite{b35}.

To impose constraints on the final herb prediction and prevent the production of herbs with side effects and duplications, we compiled the co-occurrence relationship of each symptom-herb pair from the \textit{TCM Prescription Dictionary}. We then initialized three vectors: the monarch, minister, and 'assistant and envoy' vectors, denoted as $M_c^i$, $M_d^i$, and $M_{ae}^i$, respectively. We designed the TCM-CDAE module to train the relationship between herbs and symptoms. For instance, take the monarch herb representation $M_c^i$, which we utilize as follows.
\begin{equation}
\begin{aligned}
\label{deqn_ex1a_13}
M_{\exp }=\operatorname{TCM-CDAE}_{\exp }\left(M_{c}^{i}, e_{s}\right)=\operatorname{Dropout}\left(M_{c}^{i} \circ e_{s}\right)
\end{aligned}
\end{equation}
where $\circ$ denotes the Hadamard product. For preventing drastic changes in the output neurons from affecting the model, we optimize it and obtain the ‘monarch’ herb representation $M_c$ as shown below.
\begin{equation}
\begin{aligned}
\label{deqn_ex1a_14}
M_{c}=\operatorname{TCM-CDAE}_{s q z}\left(M_{\text {exp}}\right)=\operatorname{Norm}\left(\operatorname{SumPool}\left(M_{\text {exp }}\right)\right)
\end{aligned}
\end{equation}

Eventually, we perform the following operations on the above three obtained vectors $M_c$, $M_d$ and $M_{ae}$ to satisfy the compatibility relationship between the herbs.
\begin{equation}
\begin{aligned}
\label{deqn_ex1a_5}
\left\{\begin{array}{c}\widehat{\boldsymbol{M}}_{\boldsymbol{c}}=M_{c} \\\widehat{\boldsymbol{M}}_{\boldsymbol{d}}=W_{c} M_{c}+M_{d} \\\widehat{\boldsymbol{M}}_{\boldsymbol{a e}}=W_{a e}\left(\widehat{\boldsymbol{M}}_{\boldsymbol{c}}+\widehat{\boldsymbol{M}}_{\boldsymbol{d}}\right)+M_{a e}\end{array}\right.
\end{aligned}
\end{equation}

In addition, we analyzed and compiled a total of 365 herbs from \textit{Shennong's Classic of Materia Medica}, which contains 120 'monarch' herbs, 120 'minister' herbs, and 125 'assistant \& envoy' herbs. We leverage soft K-Means to classify herbs and their associated symptom nodes into three levels based on the ‘monarch, minister, assistant and envoy’ relationship, and utilize different GCN layers to handle different levels. We focus on those herbs that are less popular nodes but have a central effect in the prescription of herbs, and the clustering method for the 'monarch' herb level is shown below.
\begin{equation}
\begin{aligned}
\label{deqn_ex1a_16}
\hat{M}_{c}^{\prime}=\sigma\left(W^{c} \cdot \sigma\left(W^{c-1} M_{c}+b^{c-1}\right)+b^{c}\right)
\end{aligned}
\end{equation}
where $\sigma(\cdot)$ denotes the Sigmoid activation function. Similar to the above method, we obtain ${\hat{M}}_d^\prime$ and ${\hat{M}}_{ae}^\prime$ for clustering.

To model the local patterns of symptoms and herbs in a fine-grained manner to support the subsequent HPR process, we fused the above features separately to obtain the final three types of herb features, where the final 'monarch' herb representation $M_C=\operatorname{CONCAT}(\hat{M}_{\boldsymbol{c}},{\hat{M}}_c^\prime)$  , the other two types of representations $M_D$ and $M_{AE}$ are similar to this process.

\subsection{Herb recommendation}

Our HPR simulates the process of clinical TCM and selects the most appropriate herb as the HPR outcome based on the summarized syndrome, which is shown below.
\begin{equation}
\begin{aligned}
\label{deqn_ex1a_17}
M_{p f}=\operatorname{ReLU}\left[\left(W_{s y n} \cdot M_{s y n}+b_{s y n}\right)\right]\left(\hat{e}_{h}\right)^{\mathrm{T}}
\end{aligned}
\end{equation}
\begin{equation}
\begin{aligned}
\label{deqn_ex1a_18}
\hat{y}_{p f_{\text {set }}}=f\left(p f_{\text {set }}, H\right)=\alpha M_{p f}+\beta\left(M_{C}+M_{D}+M_{A E}\right)
\end{aligned}
\end{equation}
where $W_{syn}$ and $b_{syn}$ denote trainable parameters.

\subsection{HPR loss optimization}

\subsubsection{Prompt CL enhanced loss}

For the PEPP module, we employ a CL framework to enhance the data for both sequences. We note that the feature elements of the patient's personalized information $x_{pf}$ are randomly masked at a rate of $\gamma_1$ to produce the augmented sequence $\hat{s_u^1} = \{\hat{I_1^u}, \ldots, \hat{I_n^u}, h_1^u, \ldots, h_{\left|PF\right|}^u\}$. Additionally, we perform zero-masking on a randomly determined portion of the herb sequence at a masking rate of $\gamma_2$, resulting in $\widehat{s_u^2} = \{I_1^u, \ldots, I_n^u, h_1^u, [mask], \ldots, h_{|PF|}^u\}$. The patient personalization loss fine-tuning loss $\mathcal{L}_p$ and the CL enhancement process $\mathcal{L}_{cl}$ are structured as follows.

\begin{equation}
\begin{aligned}
\label{deqn_ex1a_19}
\begin{aligned}\mathcal{L}_{p} & =-\sum_{\left(s, h_{i}\right) \in S_{+}^{w}} \sum_{\left(s, h_{j}\right) \in S_{-}^{w}}\sigma\left(\mathcal{U}_{s}^{T} h_{i}-\mathcal{U}_{s}^{T} h_{j}\right)\end{aligned}
\end{aligned}
\end{equation}
\begin{equation}
\begin{aligned}
\label{deqn_ex1a_20}
\begin{aligned}\mathcal{L}_{c l} & =-\sum_{B} \sum_{s \epsilon S}-\log \frac{\exp \left(\operatorname{sim}\left(u_{s}, \bar{u}_{s}\right) / \tau\right)}{\sum_{q \in S} \exp \left(\operatorname{sim}\left(u_{s}, \bar{u}_{s^{\prime}}\right) / \tau\right)}\end{aligned}
\end{aligned}
\end{equation}
where $B$ denotes the batch training size, $(u_{s}, \bar{u}_{s})$ denotes the positive examples with the embedded sequences of the original symptoms and the augmented symptoms, and $sim(\cdot)$ denotes the computation of the cosine similarity function.

\subsubsection{Loss of HPR tasks}

We define the final optimized loss $\mathcal{L}_{result}$ for TCM-HEDPR as follows.
\begin{equation}
\begin{aligned}
\label{deqn_ex1a_21}
\mathcal{L}_{\text {result }}=\mathcal{L}_{p}+\mathcal{L}_{B C E}+\lambda_{1} \mathcal{L}_{c l}+\lambda_{2} \mathcal{L}_{T C M_{-} D M}
\end{aligned}
\end{equation}

\section{Experiments and analysis}

To evaluate the TCM-HEDPR, we design a series of experiments to address these exploratory questions:

\begin{itemize}
\item{\textbf{RQ1}: How does the performance of our TCM-HEDPR compare to current more advanced HPR models?}
\item{\textbf{RQ2}: How do the key components of our TCM-HEDPR uniquely contribute to the overall performance?}
\item{\textbf{RQ3}: How does the performance of our TCM-HEDPR adapt and respond to changes key-parameter settings and does it alleviate the problem of imbalanced labels (herbs) ?}
\item{\textbf{RQ4}: To what extent does our model provide a high level of interpretability for HPR tasks, thus contributing to the decision-making process for TCM practitioners.}

\end{itemize}

\begin{table*}[h]
\caption{Overall assessment result of the model in case.}
\label{table1}
    \centering
    \resizebox{1.5\columnwidth}{!}{
    \begin{tabular}{ccc|ccc|ccc|ccc}
      \toprule[0.5mm]
      &\multirow{2}{*}{Model} & \multirow{2}{*}{Methods} & \multicolumn{3}{c}{Precision} & \multicolumn{3}{c}{Recall} & \multicolumn{3}{c}{NDCG} \\
      \cmidrule{4-12}
      & & & P@5 & P@10 & P@20 & R@5  & R@10 & R@20 & N@5 & N@10 & N@20 \\
      \midrule
      \midrule
      \multirow{4}{*}{\rotatebox{90}{\textit{ TCM\_KMGD\quad\quad}}}
      &Herb-Know & Seq2seq  & 0.526 & 0.438 & 0.343 & 0.357  & 0.446 & 0.517 & 0.362 & 0.404 & 0.512 \\
      &SMGCN & GCN  & 0.557 & 0.467 & 0.376 & 0.411  & 0.513 & 0.578 & 0.398 & 0.462 & 0.548 \\
      &KG-ASMGCN & GAT  & 0.601 & 0.519 & 0.422 & 0.474  & 0.576 & 0.612 & 0.438 & 0.588 & 0.694 \\
      &KDHR & GCN  & 0.641 & 0.562 & 0.459 & 0.518  & 0.613 & 0.759 & 0.512 & 0.664 & 0.744 \\
      &SMRGAT & GAT  & 0.676 & 0.598 & 0.511 & 0.556  & 0.670 & 0.798 & 0.552 & 0.707 & 0.784 \\
      &PresRecST & --  & 0.652 & 0.574 & 0.473 & 0.522  & 0.629 & 0.767 & 0.518 & 0.677 & 0.751  \\
      \cmidrule{2-12}
      &Ours & \textbf{Prompt \& Diffusion}  & 0.712 & 0.639 & 0.557 & 0.601  & 0.718 & 0.845 & 0.621 & 0.780 & 0.881 \\
      &\% Improv.by SMRGAT & \textbf{\& Heterogeneity network}  & \textbf{5.33\%}  & \textbf{6.86\%} & \textbf{9.00\%}& \textbf{8.09\%}  & \textbf{7.16\%} & \textbf{5.89\%} & \textbf{12.50\%} & \textbf{10.33\%} & \textbf{12.37\%}  \\
      \midrule
      \midrule
      \multirow{4}{*}{\rotatebox{90}{\textit{ TCM\_SMGCN\quad\quad}}}
      &Herb-Know & Seq2seq  & 0.306 & 0.266 & 0.225 & 0.280  & 0.379 & 0.488 & 0.351 & 0.437 & 0.527 \\
      &SMGCN & GCN  & 0.362 & 0.324 & 0.267 & 0.349  & 0.432 & 0.538 & 0.378 & 0.465 & 0.571 \\
      &KG-ASMGCN & GAT  & 0.388 & 0.340 & 0.277 & 0.398  & 0.466 & 0.574 & 0.409 & 0.503 & 0.589 \\
      &KDHR & GCN  & 0.401 & 0.366 & 0.335 & 0.447  & 0.492 & 0.613 & 0.443 & 0.532 & 0.667\\
      &SMRGAT & GAT & 0.416 & 0.381 & 0.342 & 0.479  & 0.559 & 0.667 & 0.465 & 0.584 & 0.696 \\
      &PresRecST & --   & 0.399 & 0.368 & 0.331 & 0.458  & 0.542 & 0.654 & 0.449 & 0.544 & 0.642  \\
      \cmidrule{2-12}
      &Ours & \textbf{Prompt \& Diffusion} & 0.443 & 0.409 & 0.363 & 0.499  & 0.584 & 0.691 & 0.538 & 0.643 & 0.761 \\
      &\% Improv.by SMRGAT & \textbf{\& Heterogeneity network}  & \textbf{6.49\%}  & \textbf{7.35\%} & \textbf{6.14\%} & \textbf{4.18\%}  & \textbf{4.47\%} & \textbf{3.60\%} & \textbf{15.70\%} & \textbf{10.10\%} & \textbf{9.34\%}  \\
      \midrule
      \midrule
      \multirow{4}{*}{\rotatebox{90}{\textit{TCM\_PD\_1195\quad\quad}}}
     &Herb-Know & Seq2seq  & 0.294 & 0.251 & 0.213 & 0.227  & 0.272 & 0.387 & 0.332 & 0.411 & 0.505 \\
      &SMGCN & GCN  & 0.337 & 0.287 & 0.228 & 0.258  & 0.320 & 0.446 & 0.364 & 0.452 & 0.560 \\
      &KG-ASMGCN & GAT  & 0.359 & 0.323 & 0.268 & 0.281  & 0.352 & 0.477 & 0.391 & 0.487 & 0.585 \\
      &KDHR & GCN  & 0.383 & 0.351 & 0.311 & 0.337  & 0.388 & 0.504 & 0.427 & 0.521 & 0.654\\
      &SMRGAT & GAT  & 0.408 & 0.372 & 0.334 & 0.362  & 0.453 & 0.568 & 0.464 & 0.570 & 0.681 \\
      &PresRecST & --   & 0.392 & 0.361 & 0.322 & 0.349  & 0.432 & 0.552 & 0.438 & 0.547 & 0.621  \\
      \cmidrule{2-12}
      &Ours & \textbf{Prompt \& Diffusion} & 0.434 & 0.392 & 0.354 & 0.381  & 0.472 & 0.594 & 0.514 & 0.620 & 0.744 \\
      &\% Improv.by SMRGAT & \textbf{\& Heterogeneity network}  & \textbf{6.37\%}  & \textbf{5.38\%} & \textbf{5.99\%}& \textbf{5.25\%}  & \textbf{4.19\%} & \textbf{4.58\%} & \textbf{10.78\%} & \textbf{8.77\%} & \textbf{9.25\%}  \\
      \bottomrule[0.5mm]
    \end{tabular}
    }
\end{table*}

\subsection{Experimental setup}

\subsubsection{Data sets}

To ensure the accuracy of the assessment, the following three datasets were taken: TCM\_KMGD, TCM\_SMGCN \cite{b7} and TCM\_PD\_1195 \cite{b9}. To pre-process the data, this paper utilized the 10-kernel technique to filter out the symptom-herb relationship pairs that occur less than 10 times, where: (1) TCM\_KMGD is a clinical dataset of TCM from Guangdong province cooperative hospital. Under the guidance of TCM doctors, we finally obtained 57,962 case data containing 348 symptoms and 718 herbs. In addition, we divided the patient's age by the range (the range unit takes a value equal to 5) to convert it to an age range number. (2) TCM\_SMGCN \cite{b7} is a public TCM dataset which consists of 33,765 data containing 390 symptoms and 805 herbs. (3) TCM\_PD\_1195 \cite{b9} is also a public TCM dataset which consists of 26,360 data containing 360 symptoms and 753 herbs. Since the datasets (2) and (3) lack information about patients' personalized attributes, we conduct them to evaluate the ability of TCM-HEDPR for HPR. We randomly divided this HPR task into a training set (90\%) and a test set (10\%) according to 9:1.

\subsubsection{Baselines and evaluation metrics}

To avoid errors caused by negative sampling in the evaluation \cite{b20}, we adopted Precision@K, Recall@K, and NDCG@K as the evaluation metrics for our experiments \cite{b8,b9,b10,b11}, where K stands for the number of predicted herbs.

For sequence HPR methods, we selected Herb-Know \cite{b25}, SMGCN \cite{b7}, and PreSRecST \cite{b11}. For KG-related HPR methods, we selected KG-ASMGCN \cite{b8}, KDHR \cite{b9}, and SMRGAT \cite{b10}.

\subsection{\textbf{RQ1}: Overall performance comparison}

We evaluate the overall performance of all the methods, and the results are shown in Table \ref{table1}. Based on the findings, we make the following observations:

(1) The consistent performance assessment of all methods demonstrates that our proposed TCM-HEDPR surpasses leading-edge baseline methods, underscoring its validity and superior performance.

(2) The performance evaluation results showcase the effectiveness of bootstrapping KG and DM methods compared to sequential HPR methods. The KG-related HPR methods extract fixed-dimensional embeddings from KG, restricting the learning process of herb representations and their fine-grained modeling capabilities. Moreover, these models overlook the critical symptom generalization process and herb compatibility relationships, resulting in subpar performance.

(3) While the overall performance of all models exhibits a declining trend due to the lack of patient personalization information in the two public datasets, our TCM-HEDPR still outperforms the baseline approach.

\subsection{\textbf{RQ2}: Ablation study}

We performed ablation studies on the essential modules included in the TCM-HEDPR method to demonstrate its effectiveness. For comparison with the original method, we developed five different model variants as outlined below.

\begin{itemize}
\item{\textbf{w/o PEPP}, which removed the process of pre-embedding the sequence of patient personalized attributes by the PEPP module.}
\item{\textbf{w/o DMSH}, which removed the diffusion module of the DMSH and replace this process with the variogram autoencoder, a widely used generative model.}
\item{\textbf{w/o SYN}, which removed the process of summarizing syndrome.}
\item{\textbf{w/o HGSN}, which removal of the HGSN module for herb compatibility relationships.}
\item {\textbf{w/o IKG}, which removed our manually constructed TCM knowledge graph (TCM\_IKG).}
\end{itemize}

The results of the ablation experiments presented in Table.\ref{table2}  provided significant insights into the TCM-HEDPR, leading to the following crucial conclusions:

(1) Removing the PEPP module resulted in a notable decline in performance, underscoring the importance of pre-training and embedding patient-personalized symptom profiles.

(2) The integration of the DMSH and SYN modules enhanced overall performance more effectively than others, indicating that our DM-guided methods facilitate capturing both global and local herb representation patterns, highlighting advantages in higher-order modeling.

(3) The HGSN module we developed proved entirely feasible, improving the characterization of summarized syndromes and complementing herb compatibility relationships. This addresses the overlooked issue of core herbs in a multidimensional and multilevel approach.

(4) The overall performance of the model without the TCM\_IKG demonstrated a declining trend, emphasizing the significance of the KG-guided DM method. This method aids in integrating collaborative knowledge of related entities, such as herb, into HPR training, thereby enhancing its modeling capability.

\begin{table}[!t]
\caption{Results of ablation experiments of the model in case.}
\label{table2}
    \centering
    \resizebox{1.0\columnwidth}{!}{
    \begin{tabular}{cc|ccc|ccc|ccc}
      \toprule[0.5mm]
      &\multirow{2}{*}{Model}  & \multicolumn{3}{c}{Precision} & \multicolumn{3}{c}{Recall} & \multicolumn{3}{c}{NDCG} \\
      \cmidrule{3-11}
      & & P@5 & P@10 & P@20 & R@5  & R@10 & R@20 & N@5 & N@10 & N@20 \\
      \midrule
      \midrule
      \multirow{4}{*}{\rotatebox{90}{\textit{ TCM\_KMGD\quad\quad}}}
      &TCM-HEDPR  & \underline{\textbf{0.712}} & \underline{\textbf{0.639}}  & \underline{\textbf{0.557}} & \underline{\textbf{0.601}} & \underline{\textbf{0.718}} & \underline{\textbf{0.845}} & \underline{\textbf{0.621}} & \underline{\textbf{0.780}} & \underline{\textbf{0.881}} \\
      \cmidrule{2-11}
      &w/o PEPP   & 0.684 & 0.601 & 0.518 & 0.566  & 0.661 & 0.755 & 0.606 & 0.745 & 0.832\\
      &w/o DMSH  & 0.667 & 0.584 & 0.501 & 0.545  & 0.647 & 0.736 & 0.588 & 0.724 & 0.821 \\
      &w/o SYN   & 0.691 & 0.607 & 0.524 & 0.574  & 0.676 & 0.784 & 0.611 & 0.756 & 0.844 \\
      &w/o HGSN  & 0.687 & 0.602 & 0.522 & 0.570  & 0.668 & 0.771 & 0.606 & 0.748 & 0.840 \\
      &w/o IKG   & 0.699 & 0.607 & 0.535 & 0.583  & 0.684 & 0.797 & 0.605 & 0.753 & 0.855  \\
      \bottomrule[0.5mm]
    \end{tabular}
    }
\end{table}

\subsection{\textbf{RQ3}: Further investigation on TCM-HEDPR}

\subsubsection{Sensitivity to key hyperparameters}

\begin{figure*}[htbp]
	\centering
	\begin{minipage}{0.49\linewidth}
		\centering
		\includegraphics[width=0.95\linewidth,scale=1.2]{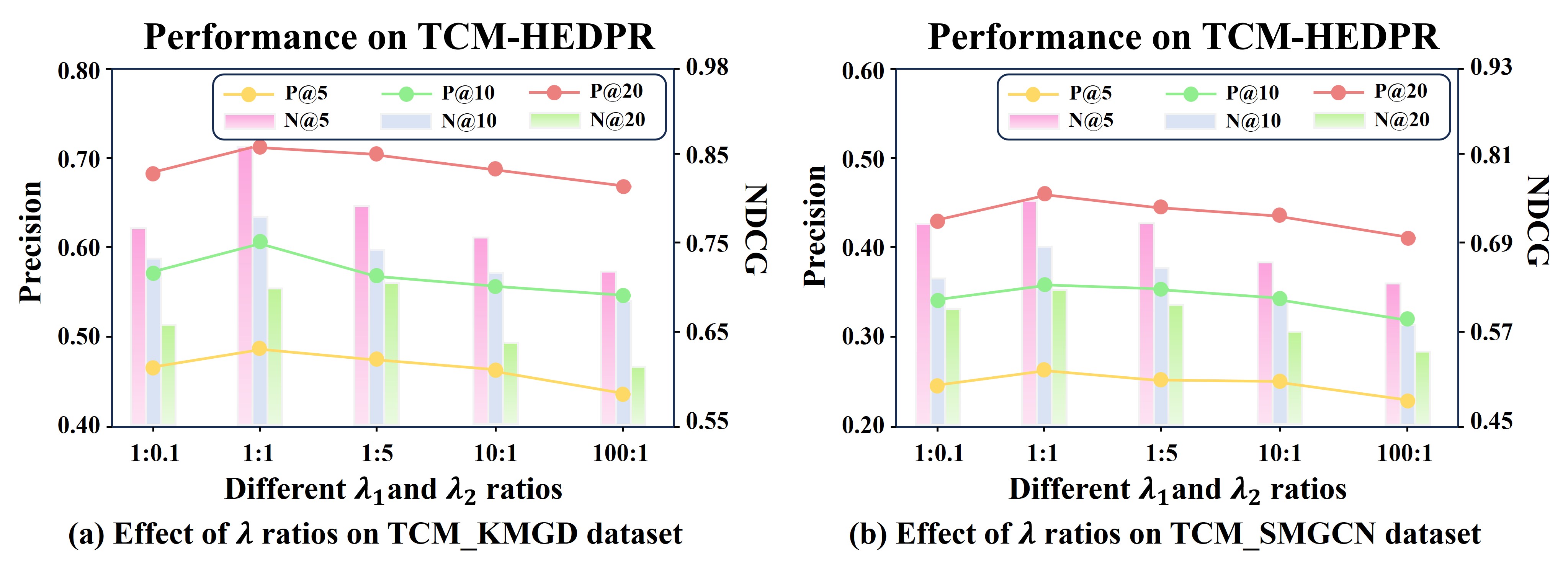}
		\caption{The effect of different $\lambda_1$ and $\lambda_2$ values for datasets.}
		\label{fig_4}
	\end{minipage}
	\begin{minipage}{0.49\linewidth}
		\centering
		\includegraphics[width=1.0\linewidth,scale=1.2]{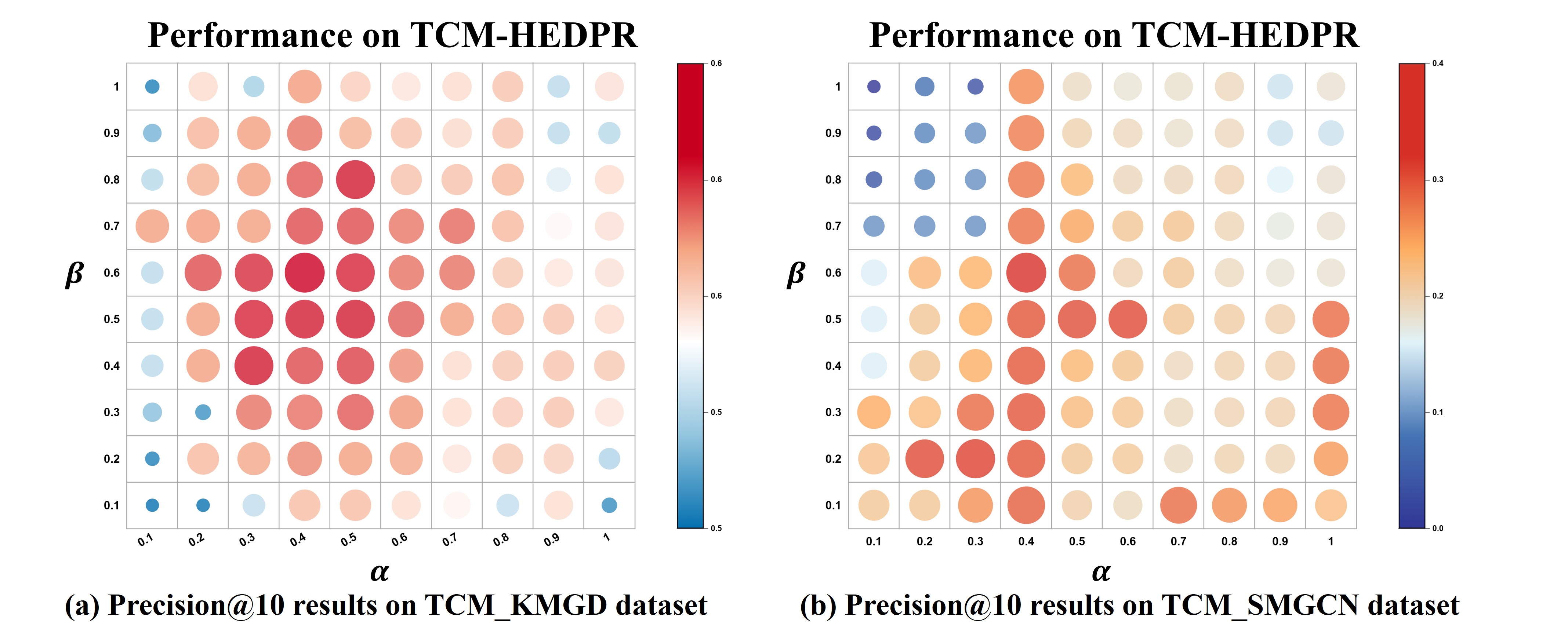}
		\caption{The effect of different $\alpha$ and $\beta$ values for datasets.}
		\label{fig_5}
	\end{minipage}
\end{figure*}

In this study, we investigate the impact of various hyperparameters on TCM-HEDPR. We employ Python 3.8.10, PyTorch 1.12.0, and NVIDIA GeForce RTX 3090 Ti GPUs. For our experiments, the embedding size is set to 128 and the batch training size to 256. For all methods, we initialize the embedding parameters using the Xavier \cite{b26} method and optimize the models with the Adam optimizer \cite{b27}. A grid search strategy is employed with a learning rate of 2e-5, a multi-head attention mechanism with \textit{N} = 2, convolutional kernel dimensions of 128, 64, and 32, and data masking rates $\gamma_1$ and $\gamma_2$ both set to 0.2. Fig. \ref{fig_4} showcased the best performance with $\lambda_1:\lambda_2=1:1$, emphasizing the significance of parameter weights. Fig. \ref{fig_5} demonstrated the relationship between the syndrome and the herbs, and the herb compatibility mechanism. We selected the optimal values of $\alpha=0.4$ and $\beta=0.6$, respectively.

\subsubsection{Imbalanced herb-labels}

\begin{figure*}[htbp]
	\centering
	\begin{minipage}{0.49\linewidth}
		\centering
		\includegraphics[width=0.9\linewidth,scale=1.2]{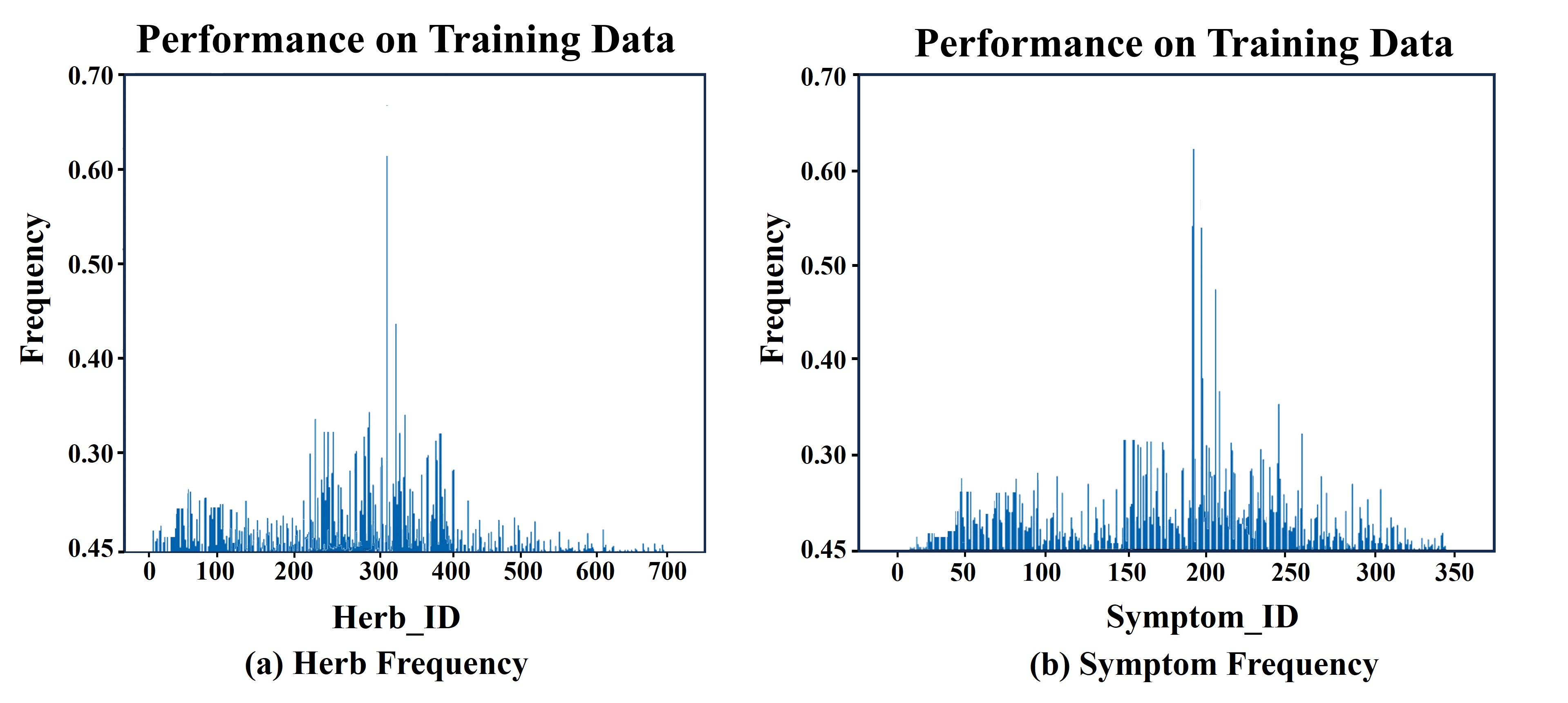}
		\caption{Frequency distribution of herbs and symptoms.}
		\label{fig_6}
	\end{minipage}
	\begin{minipage}{0.49\linewidth}
		\centering
		\includegraphics[width=0.9\linewidth,scale=0.9]{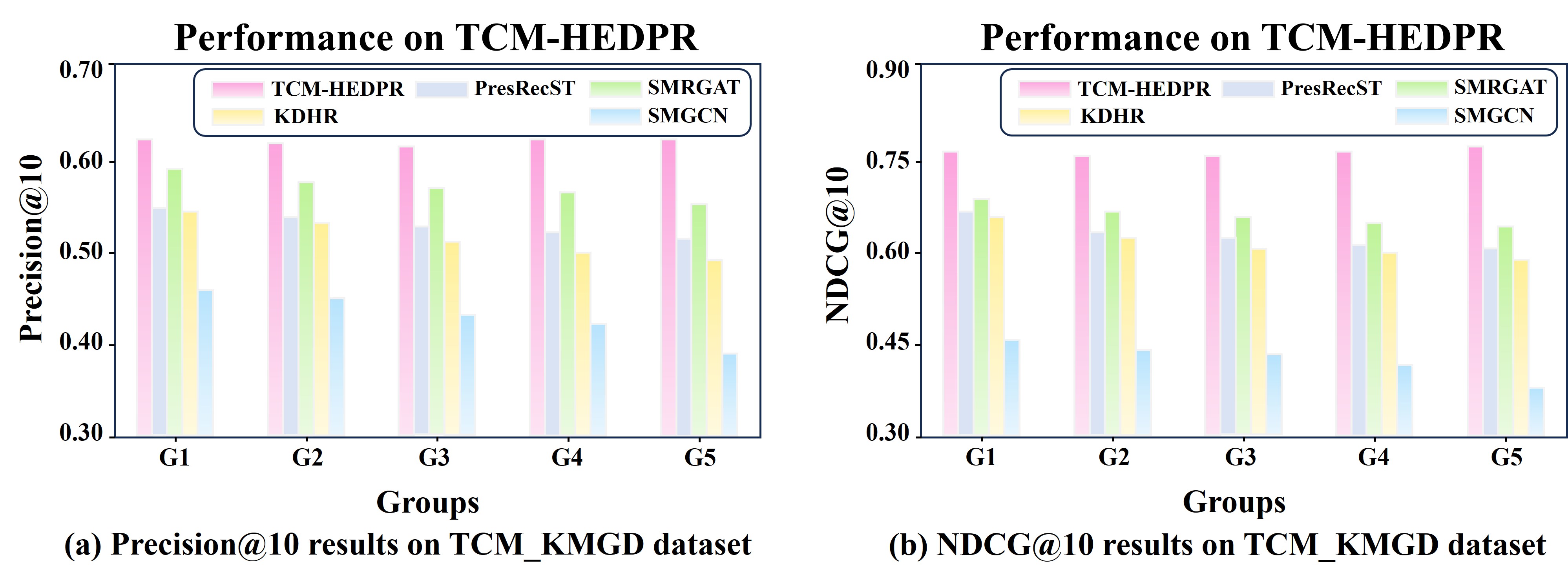}
		\caption{Sparse processing grouping of herb data.}
		\label{fig_7}
	\end{minipage}
	
	\begin{minipage}{0.49\linewidth}
		\centering
		\includegraphics[width=0.9\linewidth,scale=1.4]{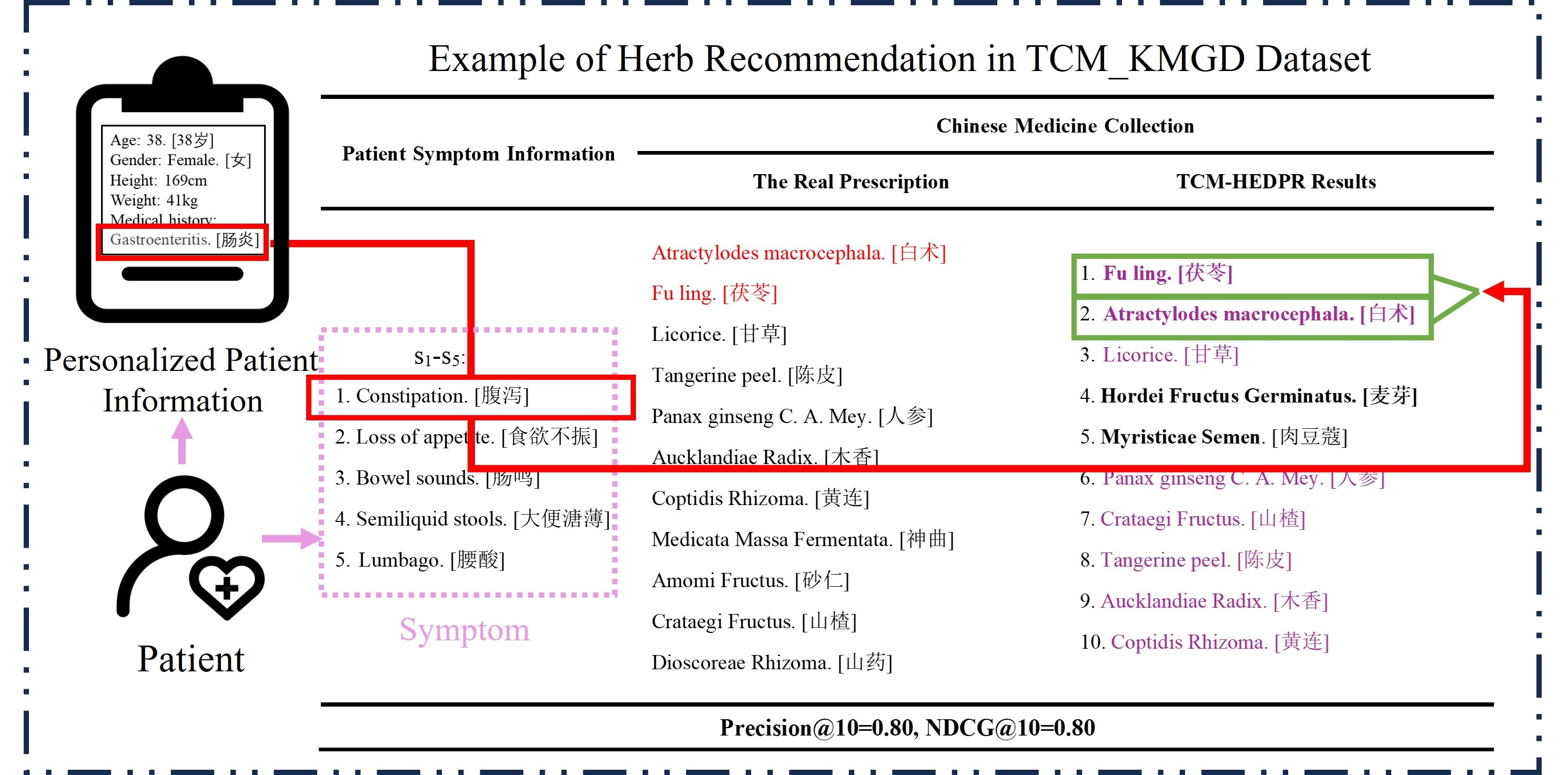}
		\caption{The herb recommendation case.}
		\label{fig_8}
	\end{minipage}
	\begin{minipage}{0.49\linewidth}
		\centering
		\includegraphics[width=0.9\linewidth,scale=1.4]{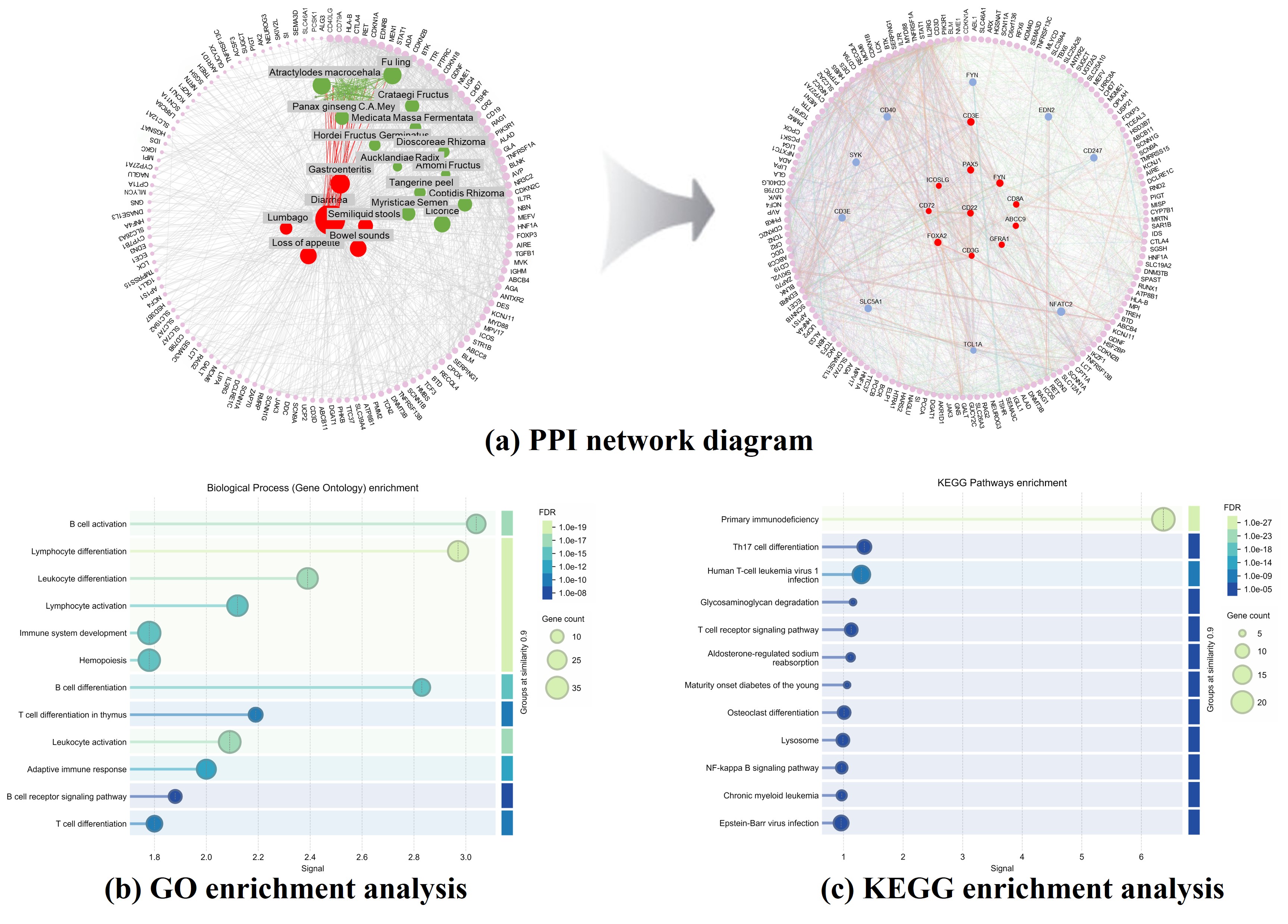}
		\caption{Network pharmacological analysis.}
		\label{fig_9}
	\end{minipage}
\end{figure*}

To evaluate our method for addressing the issue of imbalanced herb-labels, we quantified the frequency of data in the train set. Fig. \ref{fig_6} shows that the frequency of herbs exhibits a typical long-tail distribution. Therefore, we divided the herbal data into five groups, each containing the same number of herbs. The interaction density within these groups gradually increases from the first group to the fifth group, representing different degrees of sparsity. The results in Fig. \ref{fig_7} indicate that our TCM-HEDPR can significantly improve the performance of HPR. This is because baseline methods often ignore less popular TCM herbs, but sometimes these ignored herbs may belong to the ‘monarch herbs’ in TCM formulations \cite{b28}.



\begin{figure*}[htbp]
\centering
\includegraphics[width=0.84\textwidth, angle=0,scale=1.1]{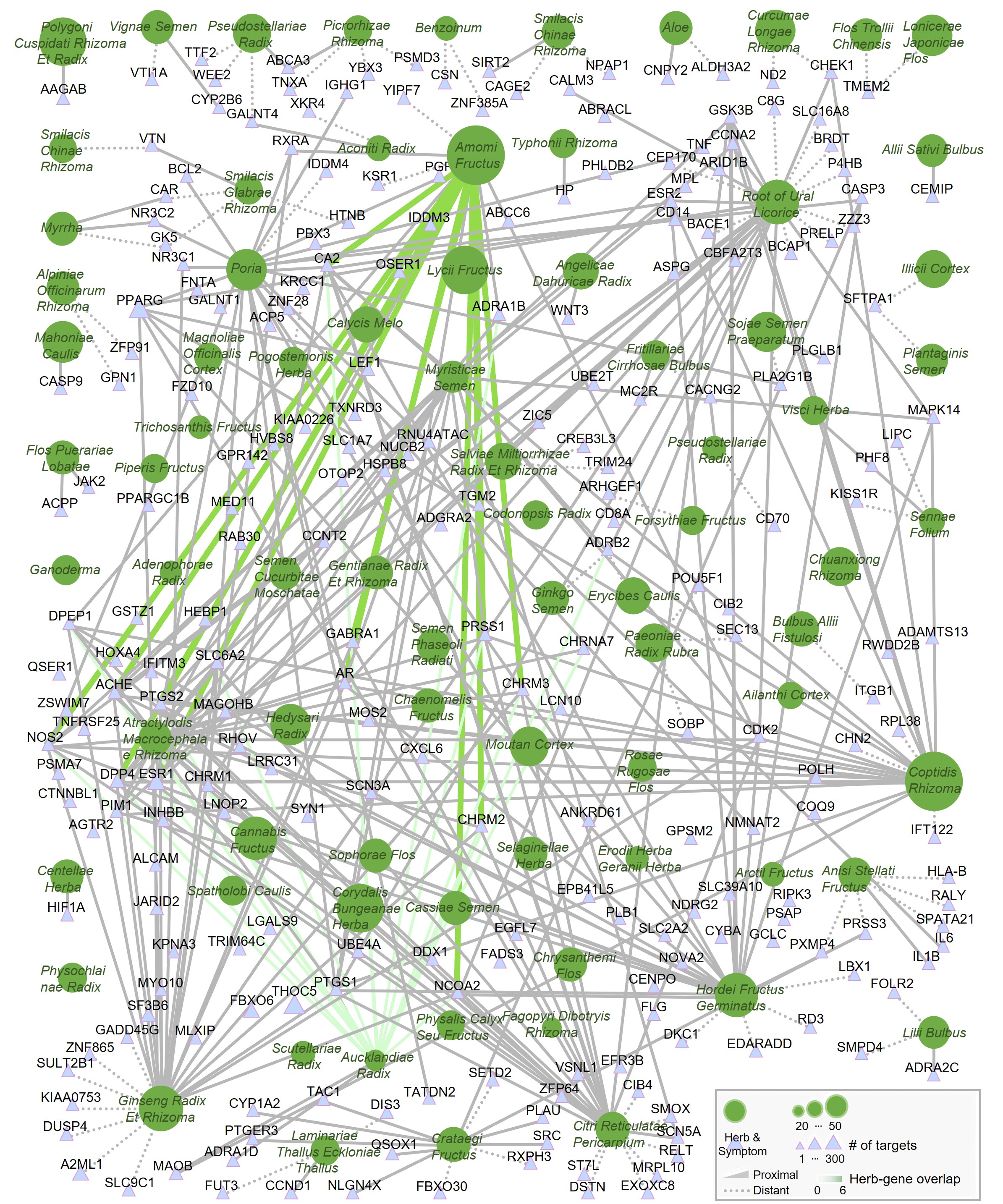}
\caption{The 'monarch, minister, assistant and envoy' targets of herb \& symptom by confidence level in case. For each known herb-symptom' target association, we connect the herb to the symptom it is used for, the link style indicating whether the herb is proximal (solid) or distant (dashed) to the disease. The line color represents the number of overlapping proteins between herb targets and symptom proteins (0, grey; 6, dark green). Node shape distinguishes targets (triangles) from herbs (circles). The node size scales with the number of proteins associated with the symptom and with the number of targets of the herb.}
\label{fig_10}
\end{figure*}

\subsection{\textbf{RQ4}: Case study}

In this section, a case study is presented to validate the feasibility of TCM-HEDPR.

\subsubsection{Herb recommendation results}

An example related to HPR is illustrated in Fig. \ref{fig_8}. Here, TCM-HEDPR recommends a corresponding set of herbs for treatment based on the patient's personalized attributes, medical history, and symptoms. Our findings indicate that the eight herbs suggested by TCM-HEDPR correspond with real prescriptions with an 80\% accuracy and hit rate (NDCG).

\subsubsection{Target interaction analysis}

To evaluate the inhibitory effects of the proposed herbal prescription's components on the target symptoms, we extracted all targets related to 6 symptoms and 13 herbs from the SymMap database \cite{b29}, as illustrated in Fig. \ref{fig_9}(a). After screening, we focused on the core symptoms 'Constipation' and 'Gastroenteritis' and their associated 161 symptom targets. Utilizing a network pharmacology approach with the STRING tool \cite{b30}, we constructed a protein-protein interaction (PPI) network to illustrate the relationships among these targets. The PPI network mainly highlights potential interference through core symptom targets like CD22 and PAX5, actual herb targets, and recommended herb targets. Node sizes indicate connectivity, with larger nodes showing interactions with numerous targets. These targets are closely linked to symptoms, suggesting a favorable therapeutic effect of the herbal combinations.

\subsubsection{GO enrichment analysis}

Fig. \ref{fig_9}(b) presents the results of GO enrichment analysis for symptomatic targets influenced by the recommended herbal ingredients, highlighting significant concentrations in 12 biological processes, including lymphocyte differentiation and the immune system. These concentrations imply that TCM-HEDPR recommended herbs regulate various biological pathways to exert therapeutic effects on symptoms. Fig. \ref{fig_9}(c) illustrates the initial results of KEGG pathway enrichment analysis and 12 signaling pathways, indicating that herbs may affect symptom targets associated with these pathways, such as primary immunodeficiency. Additionally, we examined the relationship between TCM-HEDPR-recommended herbs and actual prescription targets to elucidate the 'monarch, minister, assistant and envoy' roles in prescriptions, as shown in Fig. \ref{fig_10}. We identified the top 20 high-confidence herb targets by confidence level. Bai Shu and Fu Ling correlated with the core targets NOS2 and DPP4, whereas un-hit herbs recommended by TCM-HEDPR, such as Mai Ya and Roudou Kou, were linked to nine high-confidence core targets. In terms of herb compatibility, Mai Ya, as a 'minister' herb, collaborates with Ren Shen and others to aid in digestion and spleen strengthening (\begin{CJK*}{UTF8}{gbsn}消食健脾\end{CJK*})\cite{b31}, while Roudou Kou, associated with the symptomatic target of constipation, acts as a 'monarch \& minister' herb with Sha Ren and others for anti-diarrheal effects \cite{b32}.

\section{Conclusion}

In this study, we introduce the TCM-HEDPR model. The PEPP module is designed for the pre-embedding of patients' personalized information and data enhancement using CL. The DMSH module performs DM-guided herb representation learning, integrating the 'monarch, minister, assistant, and envoy' herb compatibility relationship provided by the HGSN module to address the issue of imbalance herb-labels. Extensive experiments across three datasets validate the effectiveness of our TCM-HEDPR approach. Ablation experiments reveal that our innovative heterogeneous graph hierarchy network and constructed knowledge graph substantially enhance the performance of TCM-HEDPR. Moreover, we examined TCM-HEDPR recommendations by applying modern medicine and network pharmacology to elucidate the transparency of our design. In future work, we plan to incorporate herb dose-related information to more accurately simulate the clinical TCM diagnosis and treatment process and to support the modernization and sustainable development of TCM.

\section*{Acknowledgments}
This work was supported by the National Natural Science Foundation of China (No.82374626).

\section*{GenAI Usage Disclosure}

No GenAI tools were used at any stage of this study.



\bibliographystyle{ACM-Reference-Format}
\balance
\bibliography{sn-bibliography}










\end{document}